\documentclass[letterpaper, 10 pt, conference]{ieeeconf}  

\IEEEoverridecommandlockouts                              
\usepackage{siunitx}            \usepackage{float}              \usepackage{amsmath}                   
\usepackage{graphicx}
\usepackage{xcolor}
\usepackage{soul}
\usepackage{lipsum}

\newcommand{\RomanNumeralCaps}[1]{\MakeUppercase{\romannumeral #1}}

\title{\LARGE \bf
ReQuBiS - Reconfigurable Quadrupedal-Bipedal Snake Robots}

\author{Harshad Zade$^{1}$, Aadesh Varude$^{1}$, Karan Pandya$^{1}$, Ajinkya Kamat$^{1}$, Shital Chiddarwar$^{1}$, \\and Rohan Thakker$^{2}$
\thanks{*This  work  was  supported  by  IvLabs  (Robotics  lab  of VNIT,  Nagpur, India)}
\thanks{$^{1}$Visvesvaraya National Institute of Technology, Nagpur, India}
\thanks{$^{2}$NASA-JPL, Caltech, Pasadena, USA.}
\thanks{(email: harshadzade09@gmail.com, adeshvarude@gmail.com, karanpandya2@gmail.com, kamat.ajinkya@gmail.com, s.chiddarwar@gmail.com, rohan.a.thakker@jpl.nasa.gov)}}
        
\begin{document}

\maketitle
\thispagestyle{empty}
\pagestyle{empty}

\begin{abstract}
The selection of mobility modes for robot navigation consists of various trade-offs.
Snake robots are ideal for traversing through constrained environments such as pipes, cluttered and rough terrain, whereas bipedal robots are more suited for structured environments such as stairs.
Finally, quadruped robots are more stable than bipeds and can carry larger payloads than snakes and bipeds but struggle to navigate soft soil, sand, ice, and constrained environments.
 A reconfigurable robot can achieve the best of all worlds.
Unfortunately, state-of-the-art reconfigurable robots rely on the rearrangement of modules through complicated mechanisms to dissemble and assemble at different places, increasing the size, weight, and power (SWaP) requirements.
We propose Reconfigurable Quadrupedal-Bipedal Snake Robots (ReQuBiS), which can transform between mobility modes without rearranging modules. 
Hence, requiring just a single modification mechanism.
Furthermore, our design allows the robot to split into two agents to perform tasks in parallel for biped and snake mobility. 
Experimental results demonstrate these mobility capabilities in snake, quadruped, and biped modes and transitions between them.
\end{abstract}
\textbf{Video:} Experimental results are available at https://youtu.be/oUigwOep0qc





\section{INTRODUCTION}\label{introduction}
\begin{figure}[t!]
    \centering
    \includegraphics[width=0.5\linewidth]{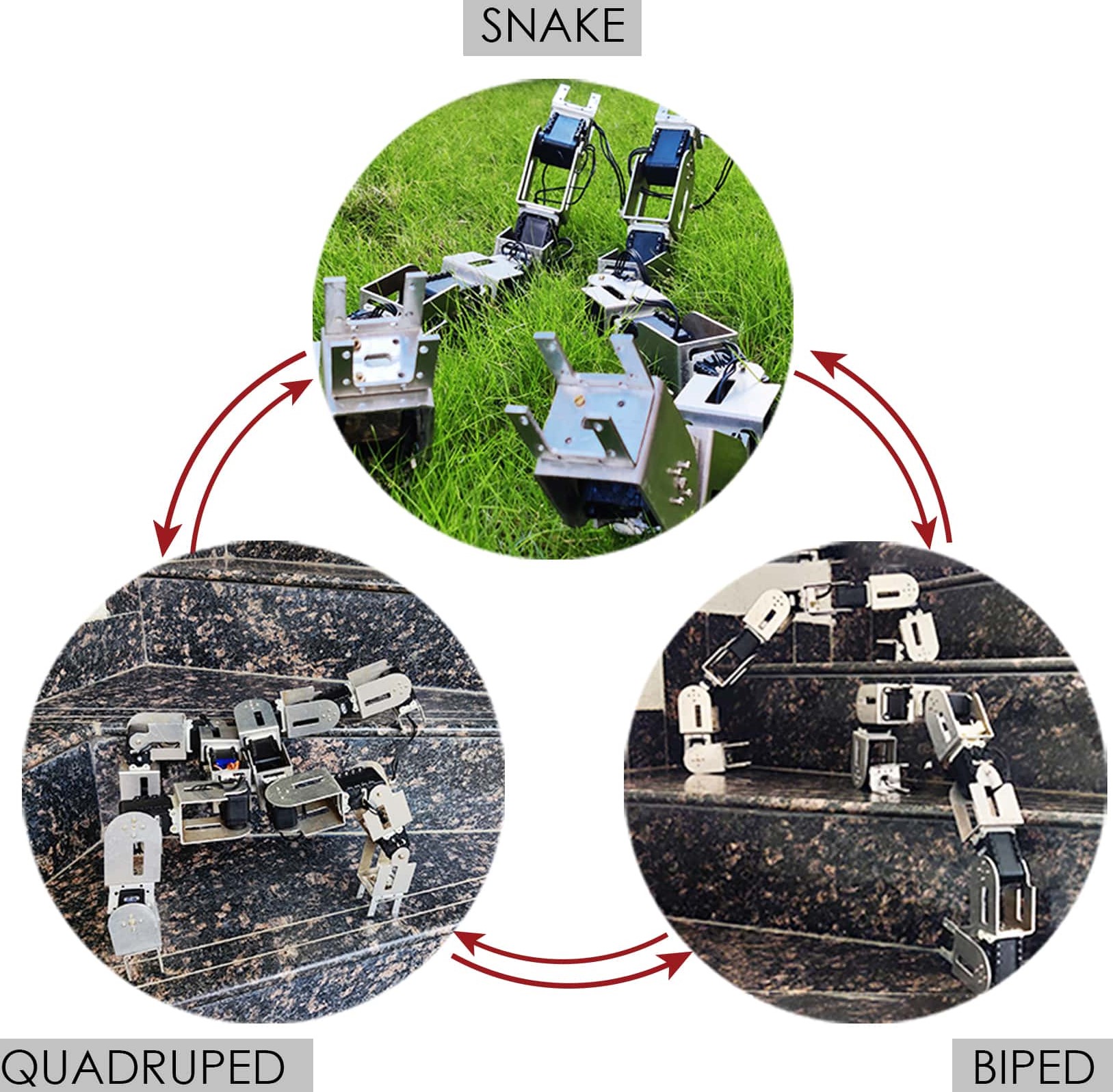}
    \caption{Illustration of various mobility modes of ReQuBiS.}
    \label{fig:State_Space_Diagram}
\end{figure}

\begin{figure}[hbt!]
    \centering
    \includegraphics[width=0.4\textwidth]{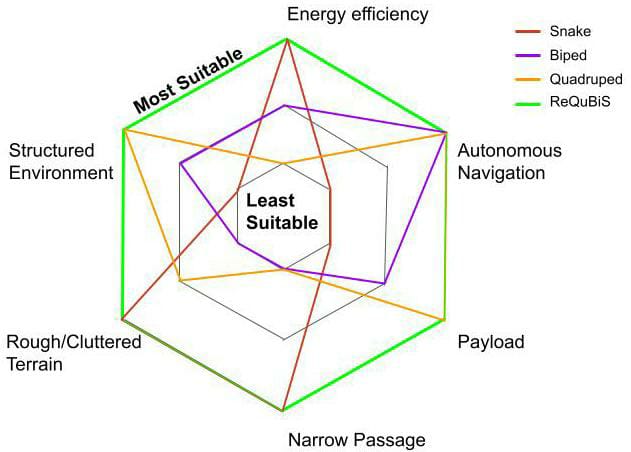}
    \caption{Comparison of strengths of various mobility modes of Snake, Biped and Quadruped with ReQuBiS.} 
    \label{fig:PCA}
\end{figure}

Robot navigation through complex terrains is an attractive open research problem with various applications in space exploration, search and rescue, etc.
Choice of mobility mode is an essential trade-off in robot design.
Fig.~\ref{fig:State_Space_Diagram} shows different mobility modes considered in this research, and Fig.~\ref{fig:PCA} provides a summary of the strengths and weakness of each mobility mode. 

Snake robots, due to their hyper redundancy and narrow cross-section of their body, are versatile and can navigate within limited space, rough terrains, move underwater \cite{c1}, traverse through pipes, slopes, and uneven ground \cite{c2}, \cite{c3}, \cite{c4}.
Snake robots can be deployed for various practical applications like locating hazardous chemical leaks, search and rescue operations \cite{c5}, reconnaissance; they can function as self-propelled inspection devices \cite{c6}, \cite{c7}, \cite{c8}. However, traversing over structured environments and carrying payloads are some limitations of snake robots \cite{c9}.


On the contrary, legged robots are helpful for traversal in structured environments such as stairs or the presence of negative obstacles by leveraging footstep planning \cite{c10}.
Also, the payload carrying capacity of legged robots, specifically quadruped, is an attractive feature \cite{c11}. 
Furthermore, compared to snake robots, odometry estimation on legged robots from proprioception is more accurate due to significantly lower slip and from exteroception is less fragile due to relatively stable movement of sensors \cite{c12}. Thus, autonomous navigation capabilities on legged systems are more superior \cite{c13}, \cite{c14}, \cite{c15}.
However, unlike snake robots, legged robots struggle while traversing through terrains with sandy or soft soil, narrow passages, or constrained environments such as pipes \cite{c16}.
Fig.~\ref{fig:PCA} sums up the above discussion.

Hence, reconfigurable robots provide a general solution that can achieve the best of all worlds.
A large class of modular robots exists \cite{c17},
\cite{c18}, but they are not designed to perform reconfiguration during the operation of the robot. 
This reduces the in situ feasibility and level of autonomy when change in mobility mode is required during operations \cite{c19}. Surveys \cite{c8}, \cite{c20} cover many reconfigurable robots developed so far. 
Some noteworthy designs include: PolyBot \cite{c21}, MTRAN-III \cite{c22}, RoomBot \cite{c23}, Superbot \cite{c24} have identical modules which attach and detach to form different configurations. 
The SMORES \cite{c25} system was developed in a way that it could replicate a majority of the above-listed robots.
Scorpio \cite{c26} tried to address the problem of changing terrains by coupling rolling and crawling locomotion gaits.


However, most of these solutions consist of many complicated modification mechanisms at different places in the robot.
This makes the system fragile and increases its SWaP.
In the case of KAIRO-\RomanNumeralCaps2 \cite{c27}, this increment in SWaP increased the torque requirements of the actuators, which forced them to use a high gear ratio, thereby reducing the velocity of the overall robot.
Furthermore, the overall transition time between mobility modes significantly increases due to the rearrangement process.

In our previous work, ReBiS \cite{c12} we demonstrated that it is possible to achieve reconfiguration between a biped and a snake without any rearrangement of modules, thus eliminating the need for complex detachment and re-attachment mechanisms called modification mechanisms henceforth.


This paper, extends this previous work by proposing a robot that can reconfigure between a snake, biped and a quadruped with a single modification mechanism. 
Furthermore, the robot can split into two agents to perform tasks in parallel in the snake and biped configurations. 
Finally, the robot can unify the agents into a single quadruped.
This allows us to use the optimal configuration for the given task and environment conditions as summarized in Fig.~\ref{fig:PCA}.

Next, we show the mechanical and electrical design of the system, followed by basic gaits for each mobility mode along with transitions. 
Finally, we demonstrate these capabilities on hardware.

\section{ROBOT DESIGN}

\begin{figure}[ht]
    \centering
    \includegraphics[width=0.5\linewidth]{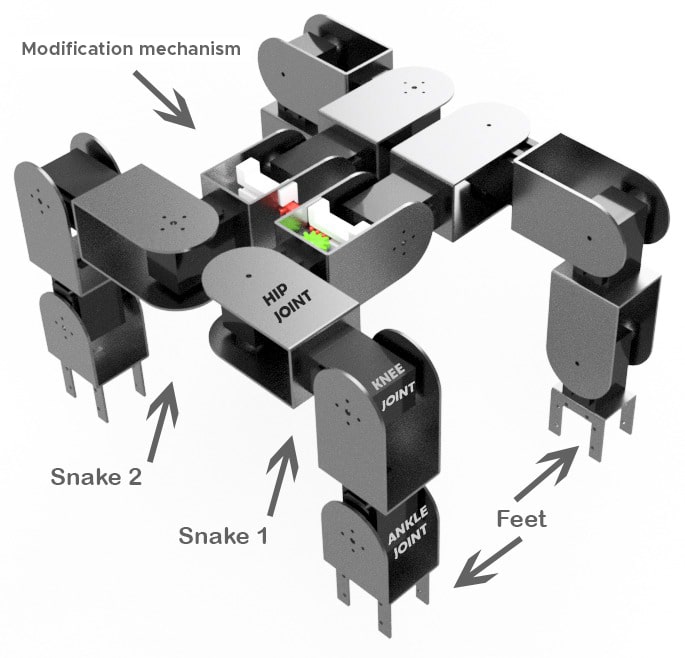}
    \caption{CAD model of quadrupedal configuration of ReQuBiS.}
    \label{fig:CAD_ReQuBiS}
\end{figure}

\subsection{\textit{Mechanical Design}}\label{MecDsg}
A ReQuBiS Snake Robot is made of 7 modular links which require higher torque motors at the extreme ends (MX-28) and comparatively lower torque motors in the middle (AX-12). 
The motors are chained together using U-links and motor clamps that are milled out of a 1.6mm thick aluminum sheet. 
Four post-like structures are attached to the terminal links that serve as the robot's feet when performing quadruped gaits.
\begin{figure}[ht]
    \centering
    \includegraphics[width=\linewidth]{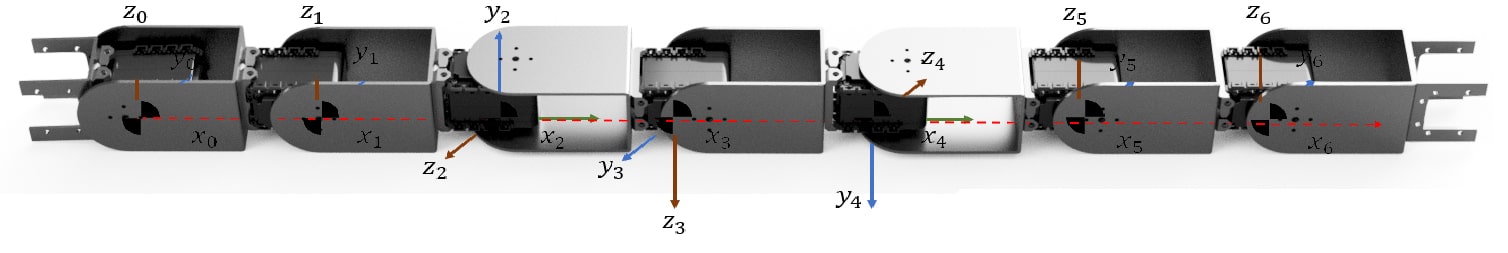}
    \caption{Frame Allocation of ReQuBiS. x(i), y(i), z(i) (where i=[0,6]) are the coordinate axes.}
    \label{fig:Frame_allocation.}
\end{figure}
Fig. \ref{fig:Frame_allocation.} shows the assembly of the robot as well as the kinematic chain that was obtained. 
To ensure motion in 3-Dimensional space, some of the links are attached at a \ang{90} offset with respect to the previous link.
It can be seen from  Fig. \ref{fig:Frame_allocation.}, that link 2 of the robot will provide rotation in the lateral plane, whereas link 3 will provide rotation in the vertical plane. 
Do note that the joints 0, 1, 5, and 6 are symmetrical and do not have any offsets. 
This allows us to use them as knee and ankle joints during quadruped locomotion.
However, this increases the length of the extreme links, and thus higher torque is required.
Therefore we used the MX-28 servo motor instead of AX-12 in the last links as they provide more torque (2.8 Nm V/S 1.2Nm). 
At the same time, the overall length of the last link, which was twice the regular length, is brought down to 1.5 times. 
Also, the contact area of those links with the ground is reduced by using the foot, as shown in Fig.\ref{fig:CAD_ReQuBiS}.

\begin{figure}[ht]
    \centering
    \includegraphics[width=0.3\textwidth]{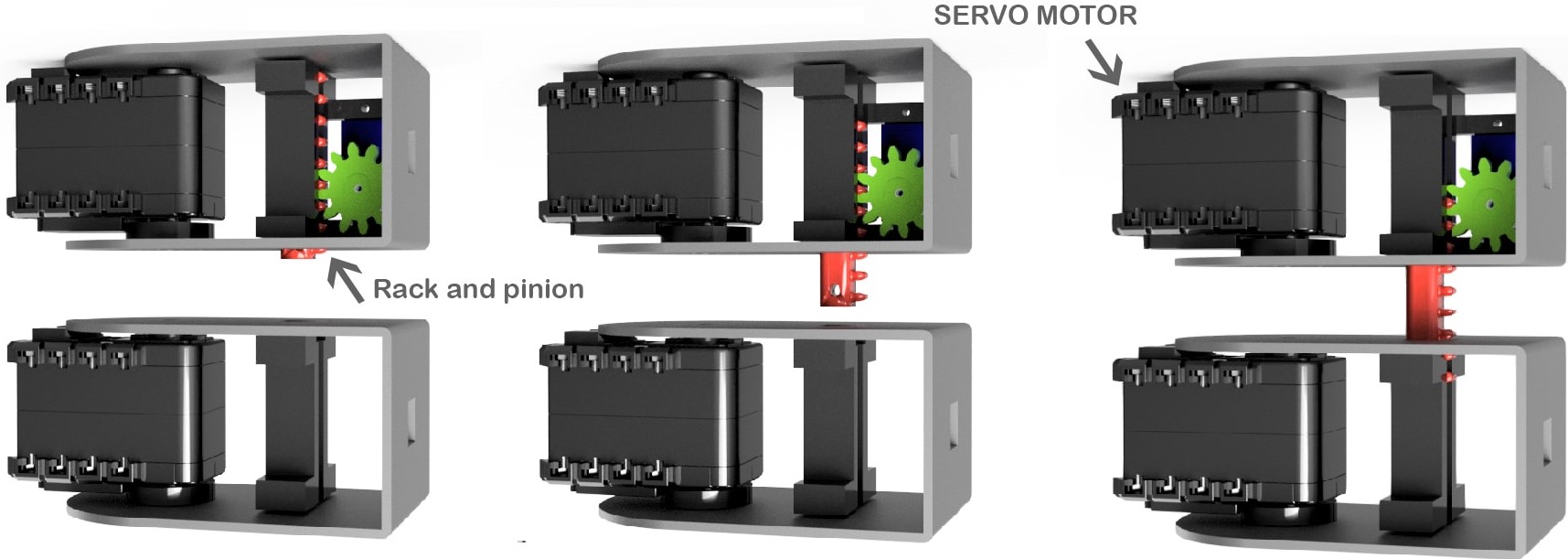}
    \caption{Modification Mechanism}
    \label{fig:Attaching Mechanism}
\end{figure}

In quadruped configuration, two ReQuBiS snakes are attached at the center links. 
The two snakes are held together with strong Neodymium magnets.
Magnets provide natural self-alignment, thus providing almost 100\% precision. 
Robustness to forces in the shear direction, is provided by a small 3D printed protrusion that is deployed by one of the robots during attachment. 
But any protrusions outside the cross section of the module would prove detrimental to snake locomotion. Hence, the attachment mechanism is designed to be completely contained within the cross-section of the module as shown in Fig.~\ref{fig:Attaching Mechanism}. 
Also, it was found that each link must have a minimum range of +90 deg to -90 deg 
to perform both snake and walking gaits. 

The former design challenge was solved by use of Neodymium magnets and a rack-and-pinion based retractable mechanism. 
The latter problems were solved by miniaturing our attachment mechanism with the help of a SG-90 micro servo, such that there is no interference between the motor and the attachment even at the full range. 
A similar companion on the adjacent link provides the additional strength required while performing the transformation gaits.

The mechanism's weight is just a fraction of the total weight (3.21\% or 43 grams).
Hence, it can accommodate heavier payloads. 
Furthermore, this mechanism helps in satisfying the aforementioned constraints. 
However, the shortcoming of this mechanism is that it requires an actuator. 
It should be noted that the modification mechanism that we propose for ReQuBiS is just one of the many possible re-attachment mechanisms \cite{c24}, \cite{c28}, \cite{c29}, electromagnetic locks.

\subsection{\textit{Electrical Design}}
ReQuBiS uses a distributed control architecture where a single microcontroller (Atmega2560) acts as a supervisor controlling 7 actuators. 
Each actuator has a closed-loop PID controller running at the joint-level that accepts desired position and velocity commands over a half-duplex TTL Bus.
To reduce wiring, the actuators are connected in a daisy chain.
The attachment mechanism was controlled using a dedicated PWM line. 
ReQuBiS is powered using a 12V off-board supply. 

\section{GAIT DESIGN AND EXPERIMENTATION}

ReQuBiS is capable of not only conventional snake gaits but also quadruped and biped gaits. 
This section describes the snake, walking, and transformation gaits we implemented on the robot.
\begin{figure}[b!]
    \centering
    \includegraphics[width=0.3\textwidth]{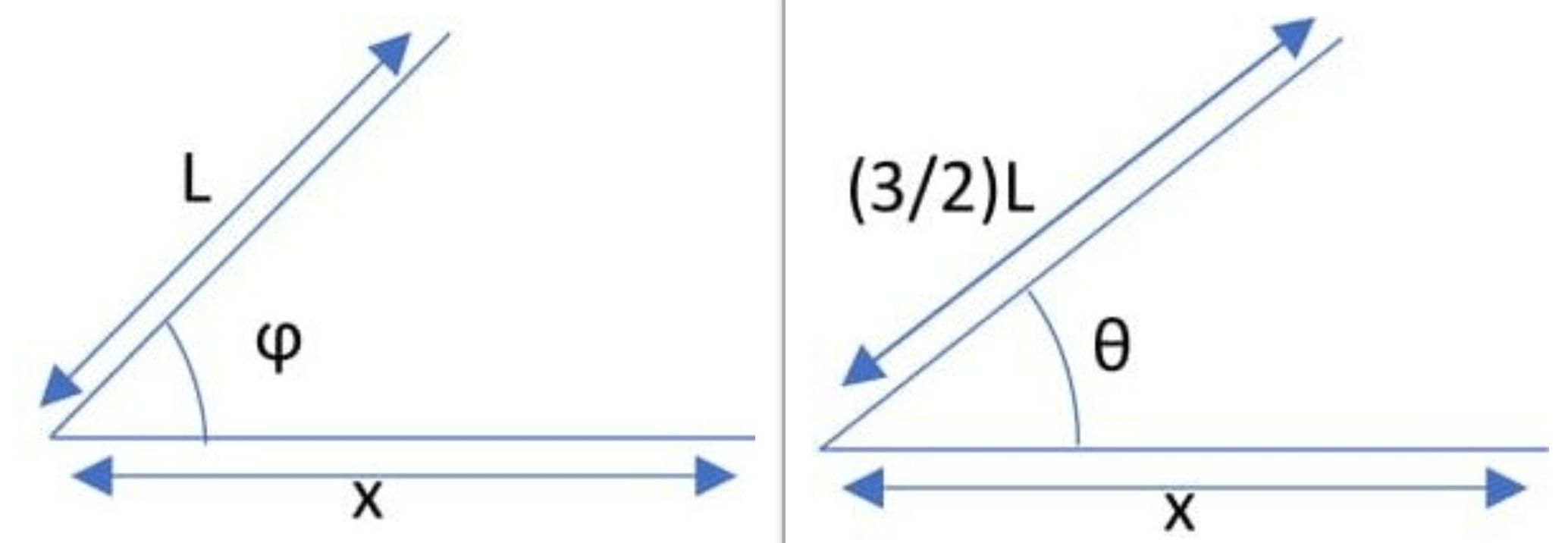}
    \caption{Different lengths and angles for extreme links.}
    \label{fig:angles}
\end{figure}
\begin{figure}[t!]
    \centering
    \includegraphics[width=0.5\textwidth]{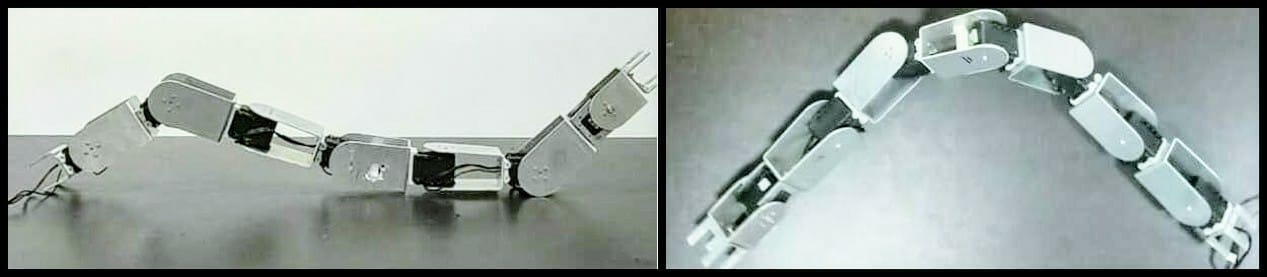}
    \caption{Snake gaits: Linear Progression (left) and Rolling (right)}
    \label{fig:Linear Progression}
\end{figure}

\subsection{Snake gaits}\label{A}
In conventional snake robots, all the joints are at an offset of 90 degrees with the previous joints. 
However, to enable walking gaits, the terminal joints have the same angle as the next joint (Fig. \ref{fig:Frame_allocation.}).
To overcome this, we fix the terminal joints to zero degrees, essentially fusing the last two links or, in other words, ReQuBiS only used 5 joints to perform snake gait. 
This, however, causes the terminal links of ReQuBiS to be 1.5 times longer than the non-terminal links.

The serpentine gaits are based on the sinusoidal curves \cite{c30} and use the standard equation for serpentine motion:
$$
\text { angle }(n, t)=\left\{\begin{array}{c}
A_x \sin \left(\omega_x  t + n \delta_x\right),\\ \text{ when } n=even. \\
A_y \sin \left(\omega_y t + n \delta_y + \phi\right),\\ \text { when } n=odd.
\end{array}\right.
$$
where,
n = the link number and
t = time.
The values of the variables $A, \omega, \delta$ are directly taken from ReBiS \cite{c12}.
ReQuBiS does not produce smooth motion following to these standard equations.
Hence, we applied a small modifier to the serpentine gait equations. 
This modifier was obtained by projecting the length of the terminal link onto the non-terminal links.

From Fig.~\ref{fig:angles}

x = length of projection of link on ground, L = length of normal link, $\varphi$ is the angle provided to the regular links, $\theta$ is the angle provided to the modified links.
\[x = \frac{3}{2}L\cos \theta\] 
\[x = L\cos \varphi\]

Now equating both the equations for obtaining $\theta$ in terms of $\varphi$:
\[ \theta=\cos^{ - 1}(\frac{2}{3}\cos\varphi)\]

\subsection{Attaching and Transforming gaits}
\subsubsection{For Quadruped}\label{AnTGait}
For the two snakes to transform into a quadruped, we assume that the location of one snake is known relative to the other. 
We start by aligning the snake robots together with high precision (below 5mm) and then engage the modification mechanism.
The rack is deployed from its nominal position i.e., completely enclosed in the module to extended position by inserting it halfway into the module of the other snake, as shown in Fig. \ref{fig:Attaching Mechanism}.

\begin{figure*}[h!]
  \centering
  \includegraphics[width=\linewidth]{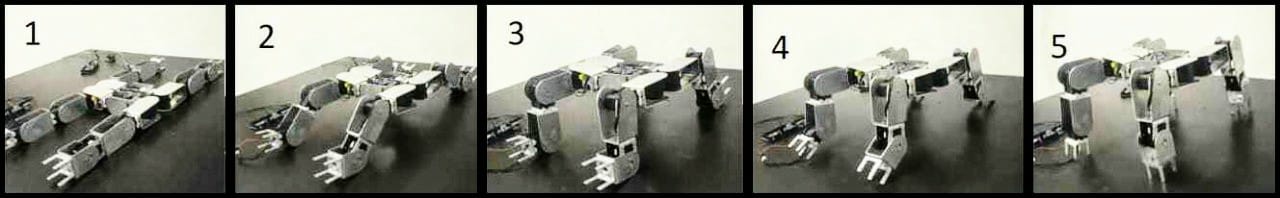}
  \caption{Transformation from snakes into Quadruped.}
  \label{fig:Transformation}
\end{figure*}

Next, we transition into the quadruped configuration by using key-frame interpolation technique, a tool we borrow from the animation community. 
Each key-frame consists of the joint position for each actuator at a particular timestamp.
As shown in Fig. \ref{fig:Transformation}, we achieve the snake to quadruped transition by interpolating through multiple key-frames.

\begin{figure*}[hbt!]
    \centering
    \includegraphics[width=\linewidth]{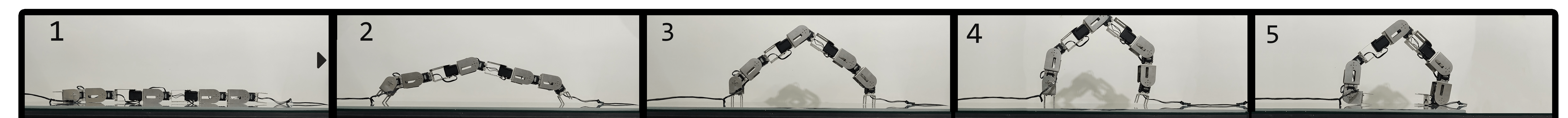}
    \caption{Snake to biped transformation.}
    \label{fig:biped_trans}
\end{figure*}

Thus, forming a quadruped robot from two snake robots.
The timing of trajectories to interpolate between key-frames is tuned to ensure dynamic stability. 


\subsubsection{For Biped}
As mentioned in section \ref{introduction}, each snake can individually transform into a biped without attaching or detaching any modules. 
The transformed biped is stable during walking as the projection of the CoM lies inside the support polygon at all instances.
The transition is again divided into different key-frames as shown in Fig. \ref{fig:biped_trans}.
The frame transition from 1 to 2 the snake form open chain straight configuration to the attained configuration by setting joint angles of extreme motors to \ang{90} and the central motor to \ang{5}. Subsequently the robot keeps rising while transitioning from one frame to another.
Since all the key-frames are in 2-dimensional space, the transition from each key-frame to the next one can be achieved using linear interpolation using the geometric approach for inverse kinematics.
In the final stage, the extreme motors are configured to \ang{90} to attain the required bipedal configuration.
 

\subsection{Walking gaits}
\subsubsection{For Quadruped}
We have implemented the static walking gait where the quadruped transformed structure is in equilibrium at every time frame of performance of gait \cite{c31}. 
Initially, the four legs of the quadruped are in contact with the ground, where equilibrium is maintained. 
At each stage of lifting the leg, the equilibrium state is maintained by ensuring that the projection of the center of mass (CoM) of the quadruped should lie in the triangle formed by the three contact points, also called as support polygon \cite{c32}, \cite{c33}.  
The gait is designed in such a way that the stability criterion is inherently satisfied. 

\begin{figure}[b!]
    \centering
    \includegraphics[width=0.2\textwidth]{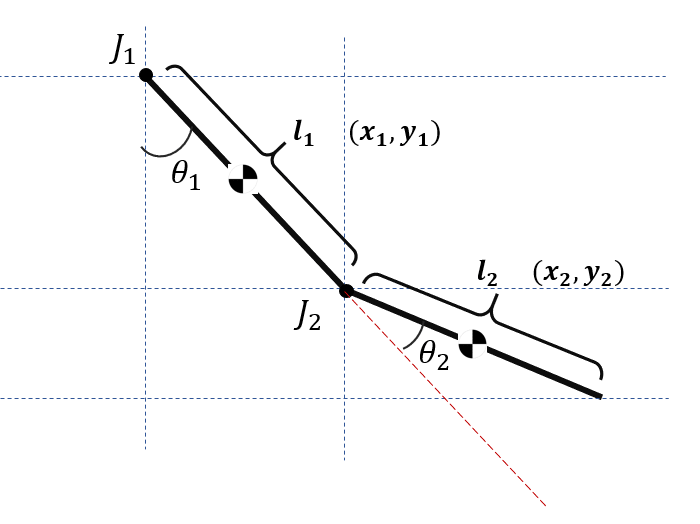}
    \caption{Double pendulum model for Quadruped leg.}
    \label{fig:leg.}
\end{figure}

For generating the gait, we assume the hip joint as the base and the remaining leg as a double pendulum as shown in Fig. \ref{fig:leg.}.
As the link frames have been established as shown in Fig. \ref{fig:Frame_allocation.}, the position and orientation of frame $i$ concerning $i-1$ can entirely be specified by four parameters known as the Denavit and Hartenberg (DH) parameters \cite{c34}.
The calculated DH parameters are given in Table \ref{dh_parameters}.
\begin{table}[h]
\caption{DH parameters of the Snake Robot Module.}
\label{dh_parameters}
\begin{center}
\begin{tabular}{|c||c|c|c|c|}
\hline
Link&$d_{i}$&$\theta_{i}$&$a_{i}$&$\alpha_{i}$\\
\hline
1&0&$0$&$112$&0\\
\hline
2&0&$0$&$75$&0\\
\hline
3&0&$90$&$75$&0\\
\hline
4&0&$90$&$75$&0\\
\hline
5&0&$90$&$75$&0\\
\hline
6&0&$90$&$75$&0\\
\hline
7&0&$0$&$112$&0\\

\hline
\end{tabular}
\end{center}
\end{table}

Following is the equation for forward kinematics:
$^{n-1}T_{n}=\left(\begin{array}{cccc}
\cos \theta_{n} & -\sin \theta_{n} \cos \alpha_{n} & \sin \theta_{n} \sin \alpha_{n} & a_{n} \cos \theta_{n} \\
\sin \theta_{n} & \cos \theta_{n} \cos \alpha_{n} & -\cos \theta_{n} \sin \alpha_{n} & a_{n} \sin \theta_{n} \\
0 & \sin \alpha_{n} & \cos \alpha_{n} & d_{n} \\
0 & 0 & 0 & 1
\end{array}\right)
$

\begin{figure}[hbt!]
    \centering
    \includegraphics[width=0.3\textwidth]{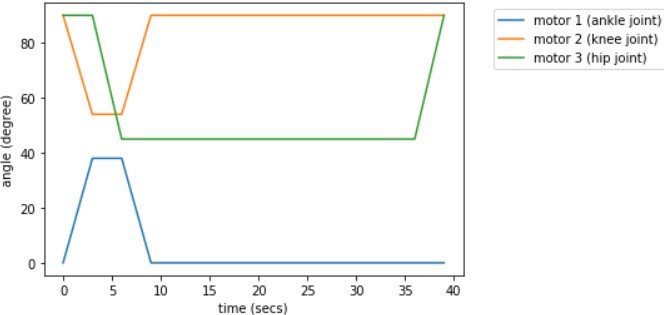}
    \caption{Joint angle trajectory for ankle, knee, hip joints of leg 1 for quadruped walking.}
    \label{fig:quad_graph}
\end{figure}

\begin{figure}[hbt!]
  \centering
  \includegraphics[width=0.4\textwidth]{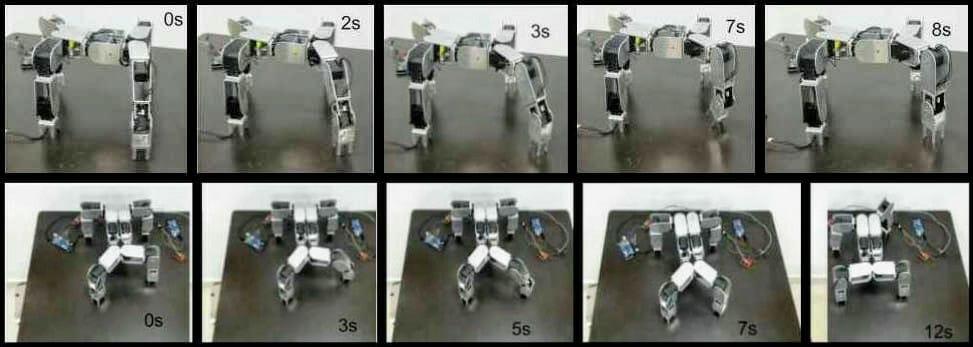}
    \caption{Walking gait of Quadruped. }
    \label{fig:Walking}
\end{figure}

Each leg lifts itself by bending the ankle, and the knee, and then the hip covers an angle of \ang{45} in the desired direction. 
Once all the legs cover an angle of \ang{45}, all the hip joints then go back to the initial position simultaneously. 
Thus, the whole body displaces in the desired direction. 
Fig. \ref{fig:quad_graph} shows the changes in angles of ankle, knee, and hip joints, with respect to time, for one leg. 
This is essentially a crawling gait that we have implemented.
Fig. \ref{fig:Walking} shows the different frames of one leg for the walking gait.
Upon running the quadruped on a graph paper, it was found that the robot traverses 11cm, on average, in one stride.

\begin{figure}[hbt!]
    \centering
    \includegraphics[width=0.2\textwidth]{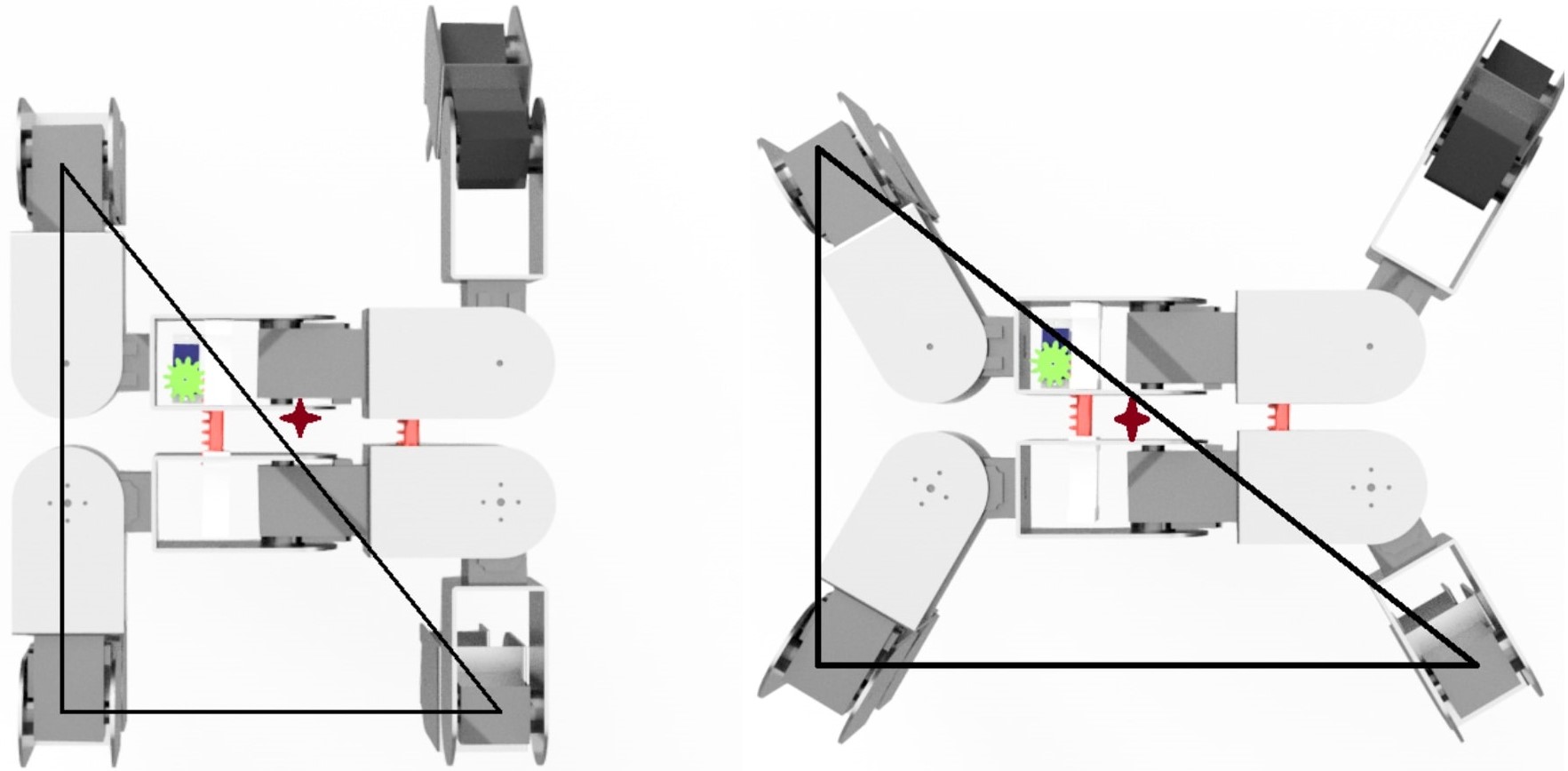}
    \caption{Comparison between area of support polygon in different configurations. Left: Lesser area. Right: More area.}
    \label{fig:Top}
\end{figure}
\begin{figure}[hbt!]
    \centering
    \includegraphics[width=0.3\textwidth]{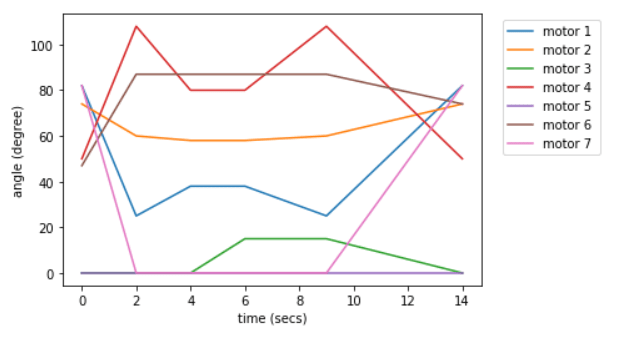}
    \caption{Joint angle trajectory for joints of 1 leg for biped walking}
    \label{fig:biped_graph}
\end{figure}
\begin{figure}[hbt!]
    \centering
    \includegraphics[width=0.5\textwidth]{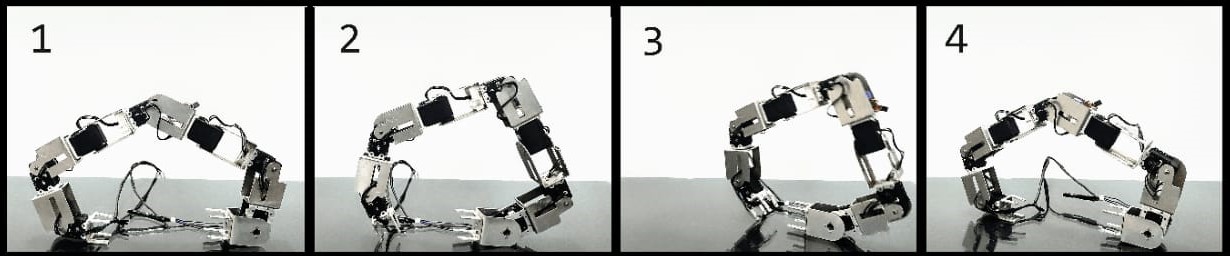}
    \caption{Walking gait for one leg of Biped. }
    \label{fig:biped_walk}
\end{figure}

On comparing the two configurations shown in Fig. \ref{fig:Top} and it can be seen that the area of support polygon is less in the left configuration resulting in the reduced stability of the structure. 
Hence the other configuration of the walking gait is employed in which the quadruped walking is creeping gait \cite{c35}\cite{c31}. 
The quadruped walks in the direction of the longer length of the body.
Here the area of support polygon is sufficient to provide stability to the structure while performing the gait, as can be seen in Fig.\ref{fig:Top} (right).
\subsubsection{For Biped}
We designed the bipedal walking gait using the concept zero-moment-point (ZMP) \cite{c36}.
In this method, first, the ZMP/CoM is shifted within the support polygon of one leg, and then the other leg is lifted and moved forward.
In biped configuration, the cross-section of the leg resting on the ground makes the support polygon.
The gait is so designed that the projection of CoM is always within the support polygon at all instances, ensuring stability. 

The robot is bent upon one leg as shown in image 2 of Fig.\ref{fig:biped_walk} to maximize the area of support polygon.
This leg acts as a base and the other leg as a manipulator of an inverted pendulum.
By using forward kinematics, it was found that the CoM lies inside the support polygon till \ang{20}.
The leg is then lifted and moved forward by an angle of \ang{15} to ensure extra safety.
After the foot is placed on the ground, the CoM is shifted on that leg.
This cycle is continued to generate the walking gait.
Fig.\ref{fig:biped_graph} shows the plot of joint angle trajectory for joints of 1 leg for biped walking.
This gait is inspired by the biped walking gait of ReBiS \cite{c12}.

\section{CONCLUSION}
In this work, we presented a reconfigurable robot that can transition into biped, quadruped, and snake configurations without any re-arrangement of modules. 
Unlike most reconfigurable robots, this key idea allowed us to achieve reconfiguration with just a single pair of modification mechanism, which significantly reduced the SWaP requirements.
The design allowed two agents to complete tasks in parallel in the snake and bipedal configuration.
These two agents combined to form the quadruped configuration, which could support payloads.
We designed and conducted a hardware demonstration of snake, biped, and quadruped locomotion gaits and transitions (refer to the supplementary material provided with the paper).
Hence asserting that ReQuBiS is capable of achieving best of all worlds by overcoming the disadvantages and leveraging the advantages of individual mobility modes as summarized in Fig.~\ref{fig:PCA}.

For future work, the design of a transition gait from biped to quadruped without going through the snake configuration is desired. 
Note that this capability is not a limitation of the hardware design and only requires gait design.
Also, exploration of more configurations, field testing/hardening and development autonomy capabilities are interesting avenues of research.


\addtolength{\textheight}{0cm}   




\section*{ACKNOWLEDGMENT}
The authors would like to thank Rajeshree Deotalu and Yogesh Phalak for meaningful mathematical discussions, which was a great learning experience. The authors are also thankful to Himanshu Mahajan, Parees Pathak, Prathamesh Ringe, Rushika Joshi for their support in fabrication and Unmesh Patil, Pranav Patil, Ujjwal Yadav, Himanshu Gorle, Atharva Sarode for their support in carrying out the experiments.

This work was done as a private venture and not in the author's capacity as an employee of the Jet Propulsion Laboratory, California Institute of Technology.


\end{document}